\documentclass[runningheads]{llncs}
\usepackage[T1]{fontenc}
\usepackage{graphicx}
\usepackage{amsmath}
\usepackage[square,sort,comma,numbers]{natbib}
\usepackage{multirow}
\usepackage{subcaption}

\begin{document}
\title{ReStory: VLM-augmentation of \\ Social Human-Robot Interaction Datasets}
\author{Fanjun Bu\orcidID{0000-0002-9953-7347} \and
Wendy Ju\orcidID{0000-0002-3119-611X}}
\authorrunning{F. Bu and W. Ju}
\institute{Cornell Tech, New York, NY, 10044, USA 
\email{\{fb266,wendyju\}@cornell.edu}}
\maketitle              % typeset the header of the contribution
\begin{abstract}
Internet-scaled datasets are a luxury for human-robot interaction (HRI) researchers, as collecting natural interaction data in the wild is time-consuming and logistically challenging. The problem is exacerbated by robots' different form factors and interaction modalities. Inspired by recent work on ethnomethodological and conversation analysis (EMCA) in the domain of HRI, we propose ReStory, a method that has the potential to augment existing in-the-wild human-robot interaction datasets leveraging Vision Language Models. While still requiring human supervision, ReStory is capable of synthesizing human-interpretable interaction scenarios in the form of storyboards. We hope our proposed approach provides HRI researchers and interaction designers with a new angle to utilizing their valuable and scarce data. 

\end{abstract}

\section{Introduction}
Compared to internet-scaled datasets used in large models, human-robot interaction (HRI) datasets in the wild are small and noisy due to the difficulty of real-world robot deployment. As a result, it might not be sufficient to train machine learning models using these datasets \cite{onHuman}. Nonetheless, such datasets are ideal for ethnomethodology researchers since they contain rich social interactions in situated contexts. In particular, human-robot communication through non-verbal movement has been a focus of ethnomethodological and conversation analysis (EMCA), where researchers examine how interactions unfold sequentially~\cite{TrashInMotion, kuzuoka_effect_2008-1, pelikan_managing_2023, pelikan2024encountering, pitsch_first_2009, tuncer_recipient_2023, yamazaki2009revealing}.

Although EMCA and machine learning are seemingly distant fields, insights from EMCA may bootstrap machine learning practices. EMCA tries to identify generalizable rules within limited, noisy data through detailed expert annotations, while machine learning algorithms leverage large datasets to detect patterns without human intervention. In HRI, even though personal idiosyncrasies are complex to model and simulate, people, as social players, tend to present themselves in predictable ways~\cite{goffman2002presentation}. When data is scarce, EMCA can point out the predictable patterns as a guide to facilitate downstream machine learning tasks. 

In our work, we augment existing HRI datasets by leveraging insights from EMCA. Since interaction datasets tend to be multi-modal, featuring both images and texts, we propose to leverage Vision Language Models to incorporate data from different modalities and sources.
The two primary challenges we face are: first, generating realistic human behaviors, and second, validating that generated interactions are appropriate in contexts. To tackle these critical issues, we introduce \textsc{ReStory}, an approach that utilizes EMCA storyboards as templates to generate human-interpretable interactions.

\section{Related Works}
\subsection{Data Augmentation for Interaction Datasets}
In domains where large datasets are difficult to access, data augmentation is an established practice to avoid model overfitting~\cite{shorten2019survey}. 
Early work that applies data augmentation in the context of HRI focuses on augmenting sensor readings during interactions rather than the interactions themselves. Murray et al. warped input audio and visual features to increase robustness in social robots' backchanneling behavior~\cite{murray2022learning}. Lakomkin et al. augmented audio features to improve robots' speech emotion recognition ability~\cite{lakomkin2018robustness}. 
More recently, foundation models have enabled the creation of new interactions driven by human engagement with these models.
For instance, Maharana et al. created the dataset LoCoMo from users' conversations with a Large Language Model-based agent, featuring long and consistent dialogues~\cite{maharana2024evaluating}.
Building upon previous work, our approach aims to investigate data augmentation approaches with a focus on behavioral interaction patterns in HRI.

\subsection{Ethnomethodology in HRI}
As robots gradually enter people's daily lives, HRI attracts attention beyond the robotics community and becomes a subject of interest in social science. Specifically, many researchers draw upon practices within ethnomethodology to investigate HRI in context \cite{jacobs2020evaluating}. One common approach is ethnomethodological and conversation analysis, which focuses on addressing how the specific interaction sequence emerges from the perspective of partakers in the situated context~\cite{pelikan2024encountering}. It focuses on the context where the interactions take place and the sequence of contingent actions at a turn-taking level. To narrate the unfolding of the interaction within context, storyboarding is the natural approach to provide readers with both spatial and temporal understanding of the scenario~\cite{truong2006storyboarding}.  
For example, Pelikan et al.~\cite{pelikan2024encountering} studied the interactions between delivery robots and passersby on public streets, where the researchers used storyboards to showcase how people interact with the robot under different circumstances. Brown et al. studied the interaction between two trash barrel robots and city dwellers in a public plaza located in New York City, where they used storyboards to illustrate the interaction patterns around the robots~\cite{TrashBarrelRobot}.

\subsection{Human Behavior Simulation}
Previous research has shown that foundation models can simulate some aspects of social behaviors and emergent interactions~\cite{aher2023using, li2024agent, park2023generative}. This capability is crucial for addressing the lack of realistic interaction training data. For instance, Park et al. utilized LLMs to simulate large-scale social interactions for social computing prototypes, aiding designers in creating better systems~\cite{park2022social}. Similarly, Duan et al. applied GPT-4 to generate human feedback on user interface mockups based on specific heuristics~\cite{duan2024generating}. Antunes et al. proposed a novel prompting technique for social scenario generation leveraging the built-in rich knowledge within LLMs~\cite{antunes2023prompting}. Given the promising interpretation and generation capability of foundation models, we aim to integrate VLMs into our pipeline for better alignment with human preferences.

\section{Introducing ReStory}
We propose \textsc{ReStory}, a data augmentation approach inspired by storyboard-based analysis commonly used in EMCA analysis. 
When showcasing an interaction on paper, ethnographers will pick key frames from the recorded video and display them sequentially in a storyboard to illustrate how the interaction unfolds. Each keyframe captures a unique message, and the readers can mentally interpolate the content among the keyframes to piece together the whole story. The keyframes represent the interaction footage in a compressed format, and adding extra frames will not improve the interpretation of the scenario. Thus, generating new interactions is equivalent to crafting a new storyboard. 

At a high level, ReStory operates as follows. Assuming an ethnographer-crafted interaction storyboard is given. We first use a VLM to caption each image in the storyboard. Then, we pick another interaction footage of a different participant, and we use VLM to caption every frame in this footage as well (for practical considerations, we choose to sample two frames per second, but this parameter is subject to change based on users' needs). Lastly, for every frame in the given storyboard, we find the image in the footage with the most similar caption to construct a new storyboard.

\begin{figure}[ht]
  \centering
\includegraphics[width=\textwidth]{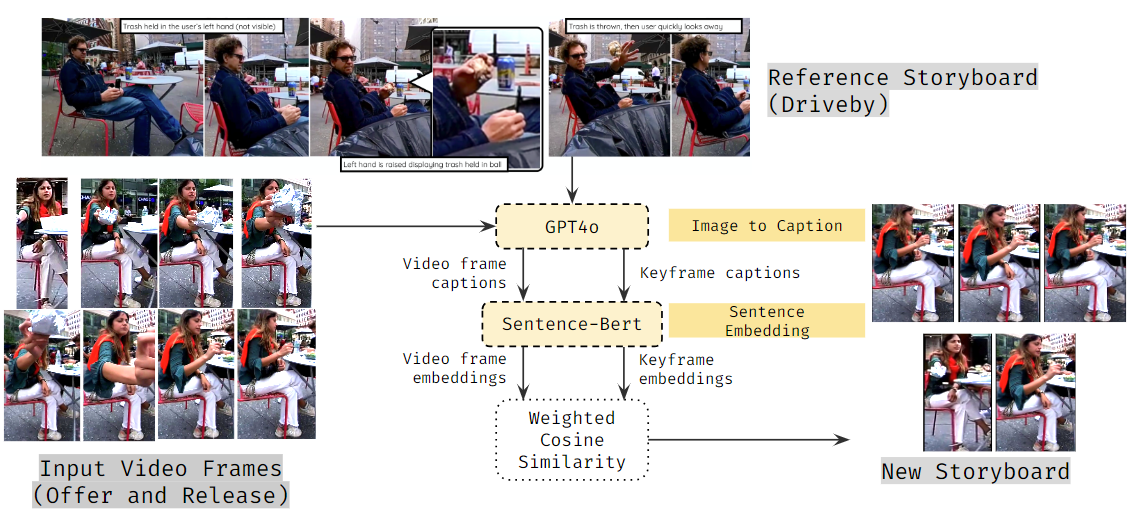}
  \caption{ReStory pipeline. The new storyboard features the woman interacting with the robot the same way the man does. The reference storyboard is from the original paper published by Brown et al. The input video is sampled at two frames per second.}
  \label{demo}
\end{figure}

To illustrate how ReStory works in practice, we took the storyboards from Brown et al. as a demonstration~\cite{TrashBarrelRobot}. Brown et al. analyzed a dataset featuring robotic trash barrels collecting trash from people in a public plaza. The robots were tele-operated by researchers in a Wizard-of-Oz way to collect interaction footage from the robot's perspective. 
In this example, as shown in Fig. \ref{demo}, we took the \textit{driveby} storyboard as the reference, where a man paid no attention to the robot until it drove by him and tossed trash inside.  For the sake of demonstration, we picked input video footage featuring a completely different interaction pattern, what Brown et al. called \textit{offer-and-release}. In this video, the woman holds a piece of trash to catch the robot's attention and baits it towards her. 

While personal idiosyncrasies may vary drastically, both the man's and the woman's behaviors can be semantically described by a sequence of atomic, repeatable actions. Due to the highly overlapping action space, we can easily rearrange the woman's actions in the order that the man executes his action, and create a storyboard of the woman interacting with the robot in the \textit{driveby} manner (see Fig. \ref{demo}). 

\subsection{Caption Model and Similarity Metric} 
The core components of ReStory are the prompts for the VLM and a task-specific metric for aligning generated captions.
Researchers can customize the prompts and measures of similarity for their intended research questions. 
In our demo, where the dataset features a mobile robot interacting with users in a garbage transaction context, we query the VLM with two prompts for every image input. The first prompt ($P_1$) focuses on the body posture of the person (e.g. \textit{"Describe the person's posture..."}), and the second prompt ($P_2$) focuses on the interaction under context (e.g.\textit{"Is the person interacting with you? How, if true, does the person interacting with you?"}). 

Since these two inquiries are highly correlated, the body poses information returned by prompting with $P_1$ is also included in the prompt body for $P_2$ (e.g., \textit{"given the person's posture is [Answer from $P_1$], is the person interacting with you?..."}). Moreover, since consecutive frames are also correlated, $P_2$ also includes the output from the $P_2$ inquiry on the previous frame (except the first frame).

For every image in both the original storyboard and input video frames, we obtain captions describing the user's pose ($C_p$) and interaction context ($C_c$). To create the storyboard, for each frame in the original storyboard, we aim to identify a corresponding frame from the input video that maximizes semantic similarity. To compute semantic similarity between sentences, we used Sentence-BERT (SBERT) models, which can take a pair of sentences as input, embed the texts into vector space, and output cosine similarity between the embeddings as a measure of semantic similarity between the input sentences~\cite{reimers2019sentence}. Specifically, for every input video frame, we define 
$$\text{Pose similarity: }Similarity_p = SBERT(C_p^{storyboard}, C_p^{frame})$$ 
$$\text{Context similarity: }Similarity_c = SBERT(C_c^{storyboard}, C_c^{frame})$$
$$\text{Weighted similarity: }Similarity_w=\alpha*Similarity_p+(1-\alpha)*Similarity_c$$
For each storyboard frame, we compute its $Similarity_w$ with every input video frame and select the input frame with the maximum $Similarity_w$ for the new storyboard. 
We weigh the context similarity over body pose similarity because we care more about what action the person is performing, either ignoring the robot or interacting with the robot, than the way they do it (i.e. throw trash with left hand or right hand). In practice, we set $\alpha$ to 0.2. 

\subsection{On Validation}
When it comes to generating human behaviors, the ultimate question is, \textit{how do we know a real person has a high probability of doing the same behavior}? In our approach, we base the new storyboard on an existing storyboard. Thus, the high-level action sequence is valid since an actual human has performed the same actions in the same order. The fact that we pull frames from another existing interaction video also guarantees that we respect the personal idiosyncrasies as well (e.g., different people gesture differently). The alignment happens only at a high-level semantic space shaped by the prompts and similarity measures. 

However, for the new storyboard to hold, there is an implicit assumption, that is, the behavior of the robot in the new storyboard must match that in the old storyboard. Every interaction involves more than one party. In the current approach, since we are generating new behavior for one party only, we must guarantee the other factors are kept constant. In the example shown in Fig. \ref{demo}, we must assume that in the new storyboard, the robot approaches the lady the same way as it approaches the man to maintain the interaction context. 

\subsection{What's Old is New}
Although we base the new storyboard on an existing interaction sequence performed by an individual in the dataset, the images are aligned using VLM embeddings at a semantic level defined by the users. Even though the high-level descriptions are similar, people's actual behavior varies drastically. For example, depending on the context of the application, a person waving with their left hand and a person waving with their right hand can both be captioned "The person is waving." However, the pixel-level information the robot sees through the sensors is drastically different. This aligns with the concept of data augmentation, where we manipulate existing datasets without collecting more data to enhance data volume and diversity. 

\section{Evaluation}
\subsection{System Validation Study}
To validate that ReStory can generate human-interpretable storyboards that capture realistic HRI, we invited seven researchers within our research group to narrate both the original and synthesized storyboards. Specifically, they were tasked to describe frame by frame what happened in the storyboard. We return to the original work by Brown et al., which shows storyboards for three different interaction patterns in the context of trash transition with a mobile robot: \textit{driveby}, \textit{offer and release}, \textit{ask and receive}~\cite{TrashInMotion}. Aside from the previously discussed \textit{driveby} and \textit{offer and release} pattern, the \textit{ask and receive} pattern features a scenario where the robot takes the initiative to approach the user and wait in front of them, "pressuring" the user to hand their trash to the robot.

We obtained the original video clips from which Brown et al. constructed their storyboard. Based on each storyboard presented by Brown et al., we utilized ReStory to synthesize two new storyboards using the other two videos that the storyboard was not originally based on. For example, for the \textit{driveby} storyboard, we synthesized two new storyboards using the \textit{offer and release}, and \textit{ask and receive} video clips. In total, we had nine storyboards, of which three were the original storyboards, and six were synthesized. 

To evaluate ReStory's capability to synthesize human-interpretable scenarios, we showed the storyboard to seven annotators. We asked them to narrate the interaction depicted in the storyboard or report where the storyboard breaks down. Since the images were taken from the robots' perspective, we provided the robots' ego-motion between each pair of consecutive frames in text format in the storyboard as well. Since the dataset owner have not publicly released the dataset, the seven annotators were selected within the research group. They had access to the video dataset but were not familiar with the footage.

\subsection{Results}
All annotators provided a reasonable narration for every storyboard, and the descriptions of atomic actions were accurate as well. However, the variation of interpretation was significant, even for the three original base storyboards.
Below, we separately discussed the annotators' narration for the three base storyboards and the six synthesized storyboards. 

For the original \textit{driveby} storyboard, four out of seven annotators described the user's initial avoidance in their narration, which was the defining signature of the \textit{driveby} pattern. Six out of seven annotators correctly described the fact that the user initiated the interaction with the robot by waving/baiting the robot in the \textit{offer and release} pattern, while the other user did not make this causality clear. Five out of seven annotators correctly described how the robot initiated the interaction with the user in the \textit{ask and receive} paradigm, while the other annotators failed to point out this relationship. However, none of the annotators described how the user was "asked" by the robot, an observation made by Brown et al. in their video analysis.

From the 14 narrations describing the two synthesized \textit{driveby} storyboards, seven capture the user's avoidance behavior. Eight narrations capture the waving/baiting characteristics of the \textit{offer and release} pattern. Six narrations capture how the robot initiates the interaction in the \textit{ask and receive} pattern. Moreover, two annotators described how the user seemed to be "asked" by the robot to provide trash, an observation that has not been made on the original \textit{ask and receive} paradigm but was discussed in Brown et al. 
We want to note that, for the storyboards that do not capture the defining characteristics of their base storyboards, the subjective narration is still valid given the sequence of images and texts.

\section{Discussion}
This paper discusses \textsc{ReStory}, an approach based on VLM to augment existing HRI datasets. The main advantage of ReStory is that it generates new interactions while respecting the interaction context. We are not advocating for any specific models being used in our demonstration.
Our main contribution is the demonstration of such a pipeline and its potential to augment HRI datasets.

\subsection{Assumptions}
ReStory relies on the assumption that at a semantic level, people's behaviors are largely constrained in an observable action space for a given interaction context. As shown in the example, people's behavior when interacting with a mobile trash barrel robot is similar across individuals (e.g., look at the robot, look away from the robot, extend arms). Naturally, people communicate with the robot through a sequence of these non-verbal signals and demonstrate their intent. Our ReStory approach pulls out these key sequences of interactions and rearranges them based on each other to generate semi-novel interactions. The preliminary system validation study demonstrates that ReStory is able to generate new storyboards that capture the semantic-level signatures of these sequences. 

\subsection{Limitation and Future Work}
\begin{flushleft}
\textbf{Perception of Distance} We notice that the perception of distance affects the annotators' interpretation of the storyboard. ReStory currently only rearranges frames based on pose and interaction context, without the understanding of distance. The misalignment of the perception of distance between the base storyboard and the generated storyboard can be one factor that drives annotators to narrate them differently. 
\end{flushleft}

For example, from the robot's perspective, a picture of a user extending their arm holding trash can be interpreted as the action of tossing if the picture is taken up close. However, if the image is taken further away, the interpretation becomes the person is baiting the robot to approach.  
ReStory currently has a hard time incorporating spatial information, which sometimes synthesizes a "toss" action that looks like an invitation (observed twice in the narrations for synthesized \textit{driveby} pattern). 
Similarly, to synthesize \textit{offer and release} pattern, ReStory sometimes selects images that were taken closer than the original baiting behavior, which gives the annotator a sense of waiting instead of inviting. 

The future version of ReStory should consider distance as an input, which can be easily computed from robots' onboard sensors or vision-based tracking and mapping algorithms. 
Moreover, if an interaction sequence only features interaction at a close distance, ReStory may need to incorporate advanced image editing functionalities to reproduce the image as taken from a distance. 

\begin{flushleft}
\textbf{Violation of Causality} Currently, ReStory does not account for causality. Given two consecutive images of a person dropping trash into the robot, where the trash was in hand in the first image and mid-air in the second image, ReStory may order the two images in the wrong order. This is expected because the current version of ReStory sees the two images as identical from a semantic level. From the six synthesized storyboards we used for the validation study, we need to swap the order of two frames in one storyboard. The future version of ReStory should account for the original frame orderings when causality matters. 
\end{flushleft}

\begin{flushleft}
\textbf{VLM hallucinations} VLMs sometimes can produce outputs that include wrong or extraneous information. As mentioned above, ReStory still relies on humans to verify the authenticity of the produced storyboards. While constraining the hallucination of VLMs itself is an active area of research, we also make an effort to improve our model's accountability. First, instead of inputting the entire frame to ReStory, we use YOLO to crop out the user from input frames, eliminating unnecessary information (Fig. \ref{demo})~\cite{yolo}. Second, we carefully design our prompts to focus on the core features relevant to the interaction context. For example, in $P_1$, we can guide the prompting by asking specifically about the user's head orientation and arm posture. 
\end{flushleft}
 
\section{Conclusion}
We propose \textsc{ReStory}, a data augmentation approach to generate semantically meaningful HRI scenarios in the form of storyboards. While the current approach still requires human supervision, ReStory is able to create human-interpretable episodes. Beyond enriching the existing interaction dataset, we believe ReStory can also be used by designers as a design tool to construct semi-novel scenarios for design ideation.
We encourage further development and refinement of our proof-of-concept approach from the community to overcome current limitations and unlock its full potential.

\section{Acknowledgement}
We appreciate Dr. Fuhao Li for his valuable discussions and insights which contributed to the ideation of this work. 

\bibliographystyle{splncs04}
\bibliography{citations}
\end{document}